\documentclass{article}

\PassOptionsToPackage{numbers, compress}{natbib}

\usepackage[preprint]{neurips_2024}

\usepackage[utf8]{inputenc} 
\usepackage[T1]{fontenc}    
\usepackage{hyperref}       
\usepackage{url}            
\usepackage{amsfonts}       
\usepackage{nicefrac}       
\usepackage{microtype}      
\usepackage{float}          

\usepackage{booktabs}     
\usepackage{tabularx}     
\usepackage{array}        
\usepackage{multirow}     
\usepackage[table]{xcolor} 
\usepackage{siunitx}

\usepackage{graphicx}
\usepackage{subcaption}
\usepackage{longtable}
\usepackage{wrapfig}

\definecolor{lightgray}{gray}{0.95}

\renewcommand{\arraystretch}{1.3} 

\title{Expertise Is What We Want}

\author{%
  Munir Al-Dajani\thanks{These authors contributed equally}\hspace{0.5em}\thanks{Correspondence: \href{mailto:ai@color.com}{\texttt{ai@color.com}}} \\
  Color Health \And
  Dr. Keegan Duchicela\footnotemark[1]\hspace{0.5em}\footnotemark[2] \\
  Color Health \And
  Kiril Kafadarov\footnotemark[1]\hspace{0.5em}\footnotemark[2] \\
  Color Health \And
  Othman Laraki\footnotemark[2] \\
  Color Health \And
  Amina Lazrak\footnotemark[1]\hspace{0.5em}\footnotemark[2] \\
  Color Health \And
  Wendy McKennon\footnotemark[1]\hspace{0.5em}\footnotemark[2] \\
  Color Health \And
  Dr. Rebecca Miksad\footnotemark[2] \\
  Color Health \And
  Jayodita Sanghvi\footnotemark[2] \\
  Color Health \AND
  Dr. Alan Ashworth\\
  UCSF \And
  Dr. Divneet Mandair\\
  UCSF \And
  Dr. Travis Zack\\
  UCSF \AND
  Dr. Allison Kurian\\
  Stanford \\
}

\begin{document}

\maketitle

\begin{abstract}
Clinical decision-making depends on expert reasoning, which is guided by standardized, evidence-based guidelines. However, translating these guidelines into automated clinical decision support systems risks inaccuracy and importantly, loss of nuance. We share an application architecture, the Large Language Expert (LLE), that combines the flexibility and power of Large Language Models (LLMs) with the interpretability, explainability, and reliability of Expert Systems. LLMs help address key challenges of Expert Systems, such as integrating and codifying knowledge, and data normalization. Conversely, an Expert System-like approach helps overcome challenges with LLMs, including hallucinations, atomic and inexpensive updates, and testability. \\

To highlight the power of the Large Language Expert (LLE) system, we built an LLE to assist with the workup of patients newly diagnosed with cancer. Timely initiation of cancer treatment is critical for optimal patient outcomes. However, increasing complexity in diagnostic recommendations has made it difficult for primary care physicians to ensure their patients have completed the necessary workup before their first visit with an oncologist. As with many real-world clinical tasks, these workups require the analysis of unstructured health records and the application of nuanced clinical decision logic. In this study, we describe the design \& evaluation of an LLE system built to rapidly identify and suggest the correct diagnostic workup. The system demonstrated a high degree of clinical-level accuracy (>95\%) and effectively addressed gaps identified in real-world data from breast and colon cancer patients at a large academic center.
\end{abstract}

\section{Background}
Oncology care, like most other domains in healthcare, relies on deep clinical expertise. After medical school, an aspiring oncologist must complete further training in internal medicine, pursue a fellowship, and then may subspecialize in a specific cancer type. However, as medicine evolves and cancer treatments become more complex, it is increasingly difficult to maintain and disseminate oncology expertise outside of large, and often urban, academic medical centers. Without distributed access to this expertise, patients may face delays in diagnosis, incomplete workups, and ultimately suboptimal treatment plans \cite{1}. \\

The National Comprehensive Cancer Care Network (NCCN) is an alliance of leading US cancer centers dedicated to defining and documenting this expertise as guidelines for cancer screening and treatment. To keep up with advancements, the NCCN frequently updates its nearly 90 guidelines across cancer types - for example, these saw more than 200 updates in a single year \cite{2}. Additionally, different institutions frequently have their own interpretation of guidelines or timelines to adopt new changes \cite{3}. Altogether, this makes it exceedingly challenging to broadly standardize care and ensure that patients are receiving care that is consistent with current best practices. \\

Beyond challenges in democratizing clinical expertise, complexities in healthcare data collection and synthesis have the potential to hinder guideline-based decision making. Today, clinical guidelines and patient data are stored in a variety of structures and formats across multiple medical record systems \cite{4}\cite{5}\cite{6}. Clinicians find themselves spending countless hours extracting relevant information from patient records \cite{7} and mapping the relevant pieces of information to the guidelines \cite{8}. Staying up to speed on best practices and implementing standards make it exceedingly challenging for clinicians to deliver guideline-informed, patient-centered care. A major contributor to this challenge is the need to interpret and transform unstructured and heterogenous patient data to standardized concepts relevant to a given workflow and the rigorous application of guideline logic to inform clinical decision-making. We demonstrate an LLM architecture that efficiently and accurately helps deliver a critical task in cancer care – identifying workup gaps prior to starting cancer treatment. This architecture is naturally extensible to similar problems in guideline-based medicine for cancer care and other medical disciplines. \\

\section{Problem Overview}
While extensive investments have been made in building tools to support clinical decision making, these tools have been both challenging to build and impractical for clinicians to use. We break down these challenges into two broad categories. The first revolves around the accurate and efficient encoding of clinical guidelines in software in a manner that is consistent with actual clinical practice. The second category of challenges is in the practical implementation of these tools in a clinical setting.

\subsection{Challenges Representing Clinical Guidelines in Software}

Two common strategies for representing clinical guidelines in software are Machine Learning (ML) and rule-based (RB) strategies. ML systems such as LLMs build statistical models from data to generate probabilistic predictions, enabling them to encode complex relationships. RB systems such as Expert Systems rely on explicitly defined rules to create a direct software representation of the logic for a given domain. Each of these strategies on its own runs into substantial challenges when used to represent clinical guidelines, these are described in Table~\ref{tab:ml-rb-challenges}. 

\begin{table}[htbp] 
\centering
\caption{Challenge areas for representing clinical guidelines with ML \& RB systems}
\label{tab:ml-rb-challenges}

\begin{tabularx}{\textwidth}{%
  p{0.35\textwidth}                         
  >{\centering\arraybackslash}m{0.5cm}      
  X                                         
}
\midrule

\multirow{2}{*}{%
  \parbox{0.33\textwidth}{%
    \textbf{Translating Logic}\\[4pt]       
    \emph{\small                            
      Clinical guidelines are written for human consumption by medical professionals, 
but implicitly embed assumptions about domain knowledge and include contradictions, ambiguities, and gaps.
    }
  }
}
& \textbf{ML}
& Training a model to encode guideline logic based on 
real-world data incorporates practical errors, biases, engenders privacy risk, and reliance on large / representative datasets to comprehensively cover the problem space. 
\\
\cmidrule(lr){2-3} 
& \textbf{RB}
& Translating rules from clinical guidelines into software is 
a slow and error-prone process. Minor misunderstandings can result in many cycles spent debugging and trying to explain system behavior.
\\
\midrule

\multirow{2}{*}{%
  \parbox{0.33\textwidth}{%
    \textbf{Updates}\\[4pt]
    \emph{\small
      As our understanding of medicine evolves, so do dictionaries of terminology, clinical decision criteria, and available treatment options. 
    }
  }
}
& \textbf{ML}
& Achieving logic updates and revalidation via training data inherently makes changes onerous and risky. Adapting a model to changes in the underlying logic 
it’s meant to represent requires significant updates to the training data. Training data that encodes outdated concepts needs to be identified and scrubbed, and new training data needs to be introduced to encode new concepts.
\\
\cmidrule(lr){2-3} 
& \textbf{RB}
& As a rule-based system evolves, the growth of rules required to represent a complex domain leads to systems that are difficult to manage and maintain. The interdependencies between rules create cascading challenges when updates are made, requiring extensive revalidation to prevent unintended errors \cite{9}.
\\
\midrule

\multirow{2}{*}{%
  \parbox{0.33\textwidth}{%
    \textbf{Customization}\\[4pt]
    \emph{\small
      Clinical guidelines do not always converge to a global consensus. Different institutions or individual experts often have different perspectives on what the best practice is for a particular case or may be subject to different constraints/priorities. While most institutions agree on a common foundation, the ability to support custom logic robustly and cost-effectively is essential. 
    }
  }
}
& \textbf{ML}
& Two suboptimal options are available. The first is for each institution to re-train or fine tune a model. This runs into the same challenges as updates. The second is to incorporate custom logic at runtime (eg. RAG), but this also proves to be difficult to do robustly - for example to map internal model representations to the institutional model.
\\
\cmidrule(lr){2-3} 
& \textbf{RB}
& Implementing changes on top of a rule-based system is challenging for reasons described in how these systems handle updates; introducing risk and necessitating costly software engineering for every deployment.
\\
\midrule

\multirow{2}{*}{%
  \parbox{0.33\textwidth}{%
    \textbf{Testing \& Validation}\\[4pt]
    \emph{\small
      Clinical guidelines are required to interface with a complex, non-standardized world. When these are encoded into software, testing and validating that a system behaves correctly across all of the possible edge cases that may be encountered is both important and challenging.
    }
  }
}
& \textbf{ML}
& Machine learning models make testing and validation challenging for two reasons. First, their stochastic nature makes outputs unpredictable and inconsistent - unlike traditional computer code,  reproducibility is not guaranteed. Behaving correctly in one instance does not guarantee future correctness with the same or equivalent output. Second, machine learning based models generally operate as black boxes, making it challenging to have confidence that edge cases do not cause unpredictable or inconsistent behaviors.
\\
\cmidrule(lr){2-3} 
& \textbf{RB}
& Strict rule-based systems require constrained inputs in order to operate. While the software-only component is testable, the reality of real-world clinical data requires the system as a whole to rely on human translation of unstructured inputs. The reliance on human translated inputs makes testing and validation expensive and difficult.
\\

\bottomrule
\end{tabularx}
\end{table}

\subsection{Challenges in Clinical Implementation}

In addition to the challenges of representing clinical guidelines and logic in a software-based system, a number of clinical implementation challenges need to be addressed in order for a tool to be usable and trusted by physicians.

\paragraph{Data quality} Medical data is often complex, incomplete, stored in inconsistent formats, and represented as unstructured text (e.g. clinical notes). Additionally, patient data is frequently distributed across multiple institutions and transferred through PDFs and scanned images without any common data format. This requires clinicians to laboriously pore through documents to extract key information necessary for clinical decisions. As a result, clinical decision support tools can be difficult to use, expensive to integrate, or limited in their applicability \cite{10}.

\paragraph{Integrating clinical judgement} Virtually no patient’s attributes perfectly fit the guideline structure. For tools to be effective, they require the ability to apply clinical judgement across a variety of circumstances such as incomplete data, unique patient circumstance / preference, ambiguous or non-conclusive consensus in guidelines. Both black box machine learning and rigid rule based systems make it difficult to offer the flexibility needed to capture the breadth and scope of patient cases seen within standard clinical practice.

\paragraph{Explainability} To ensure clinician trust and maintain patient safety, a clinical decision support tool must be able to provide the reasoning behind an output. LLM based tools are vulnerable to hallucinations that offer plausible, yet incorrect explanations \cite{11}. On the other hand, rule-based systems lose important information when mapping the context-rich information in health records to the discrete structure with which they encode logic and evaluate decisions. 

\section{Color's Approach}
\subsection{A Large Language Expert Architecture}

Color designed a novel architecture, the Large Language Expert (LLE), that applies complex and dynamic clinical guidelines to real-world medical data. This architecture is a hybrid LLM \& Rule-Based system that leverages the strengths of each approach to overcome the limitations of the other. The LLE powers applications that flexibly interface with unstructured data while maintaining a robust, declarative logic model to drive decisions.

\begin{figure}[htbp]
    \centering
    \includegraphics[width=\linewidth]{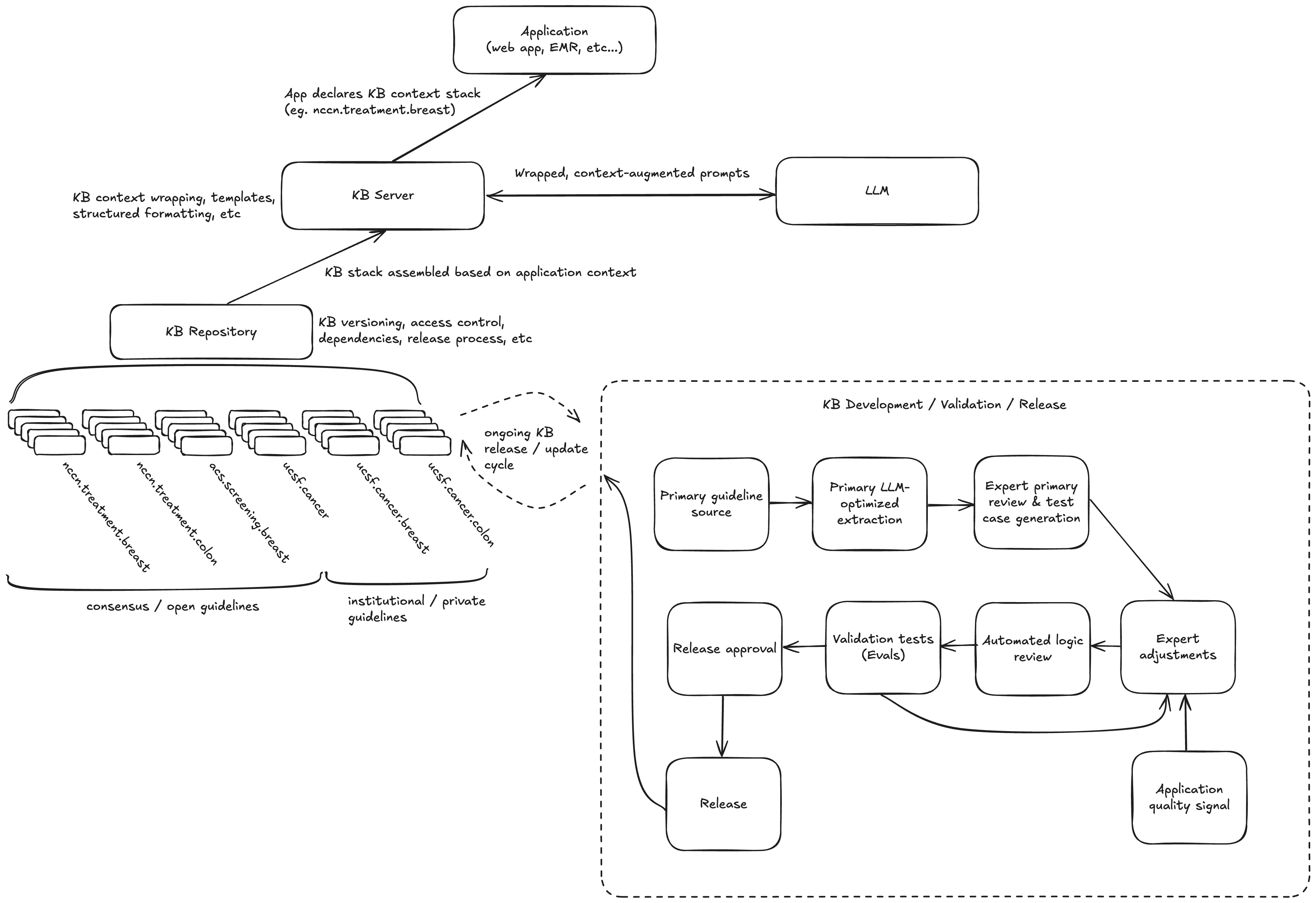}
    \caption{Application Architecture with Large Language Expert}
    \label{fig:lle_arch}
\end{figure}

The LLE evolves ideas behind Retrieval Augmented Generation (RAG) models in the direction of structured expert system formal models. In the LLE architecture, guidelines are organized and deployed as knowledge bases composed of natural language and structured logic that are namespaced and versioned. At runtime, the application invokes one or multiple knowledge bases for a given workflow. While this adds rigidity to the knowledge base content and format, it enables robust interrogation of reasoning, consistency, and biases in using these systems in the real world. The integration of this LLE system into an application is shown in Figure~\ref{fig:lle_arch}.

With this architecture, it becomes possible to:
\begin{itemize}
    \itemsep0em 
    \item check the logic for consistency 
    \item build robust test cases 
    \item build robust pipelines to transform the primary guideline sources into a knowledge base format
    \item build applications that expose the logic as a human expert would reason about a problem
    \item make deterministic/declarative logic updates
\end{itemize}

\subsection{Knowledge Base Development}
At the core of the LLE model is the translation of clinical guidelines that codify expert-guided workflows. The LLE approach is to translate specific, versioned expert documents into an LLM-optimized declarative format. The artifacts generated from this translation step remain human-readable, which enables domain expert review and adjustments. However, the extraction of a formal logic structure makes it possible to automate logic checks, identify gaps and contradictions - and most importantly power robust and reliable workflows. By using structured knowledge bases, we are able to generate targeted and streamlined LLM prompts which yield highly accurate results.

\begin{figure}[htbp]
    \centering
    \includegraphics[width=0.85\linewidth]{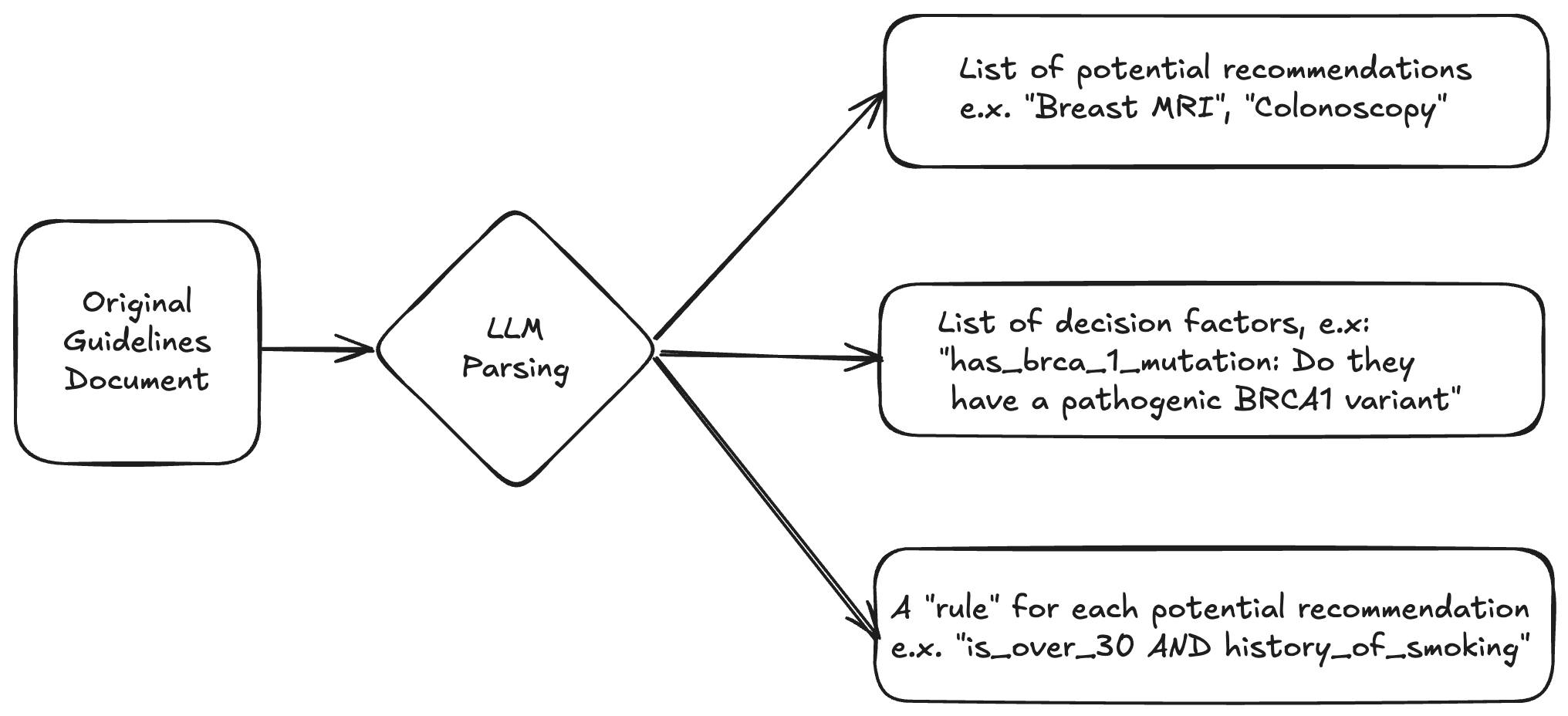}
    \caption{Extraction of Decision Factors \& Rules from Guidelines}
    \label{fig:lle_parsing}
\end{figure}

\paragraph{Translating Logic}
An obstacle in managing expert system knowledge bases is the difficulty converting a complex, interweaved, and interdependent set of expert rules into executable code. In the LLE, we bypass this complexity by staying as close as possible to the original guidelines and using English as the “executable code.” The advanced reasoning capabilities of OpenAI’s o1 model \cite{12} enable us to accurately parse the logic embedded within natural language in clinical guidelines and translate it to first-order logic which undergoes an expert review process to ensure accuracy. Through this process, we extract from the document the possible clinical recommendations, the decision factors that are used to determine what to recommend, and a rule for when to recommend them, as shown in Figure~\ref{fig:lle_parsing}. Translating the guidelines into first-order logic enables the identification of contradictions, ambiguities, and gaps, allowing experts toeasily review and resolve them.

As reasoning capabilities of LLMs improve, this process will continue to see improvement in streamlining the knowledge base generation process. We have seen that as we have adopted newer and better models, starting with GPT-4Turbo until o1, the quality of the zero-shot generated first-order logic has progressively required less expert intervention. This significantly reduces the cost to introduce new knowledge bases or update existing ones. 

\paragraph{Updates \& Versioning}
An update to a clinical guideline involves a simple change in natural language in a knowledge base. The corresponding update in the LLE is done independently, with experts modifying only the affected rules. With this approach, our update cycle is quick and cheap. Further, since the rule change is done by the experts themselves, there is less room for semantic changes to be lost in translation when implemented. Once the update is made, the translated guidelines are snapshotted into a versioned artifact that the application can configuratively invoke.

Additionally, the ability to maintain a versioned history of released knowledge bases can support important practical use cases that become challenging in a single Machine Learning architecture. For example, while guidelines are frequently updated, institutions adopt these changes at different times and determine their own schedules for staff training and protocol implementation. Another example of the value of versioning is for audit and quality measurement purposes. A health system may want to post-process its records to assess the level of guideline compliance and understand circumstances that lead to deviations from guidelines. Naturally, that would need to be done with the protocol versions used during the period being reviewed.

\begin{figure}[htbp]
    \centering
    \includegraphics[width=0.85\linewidth]{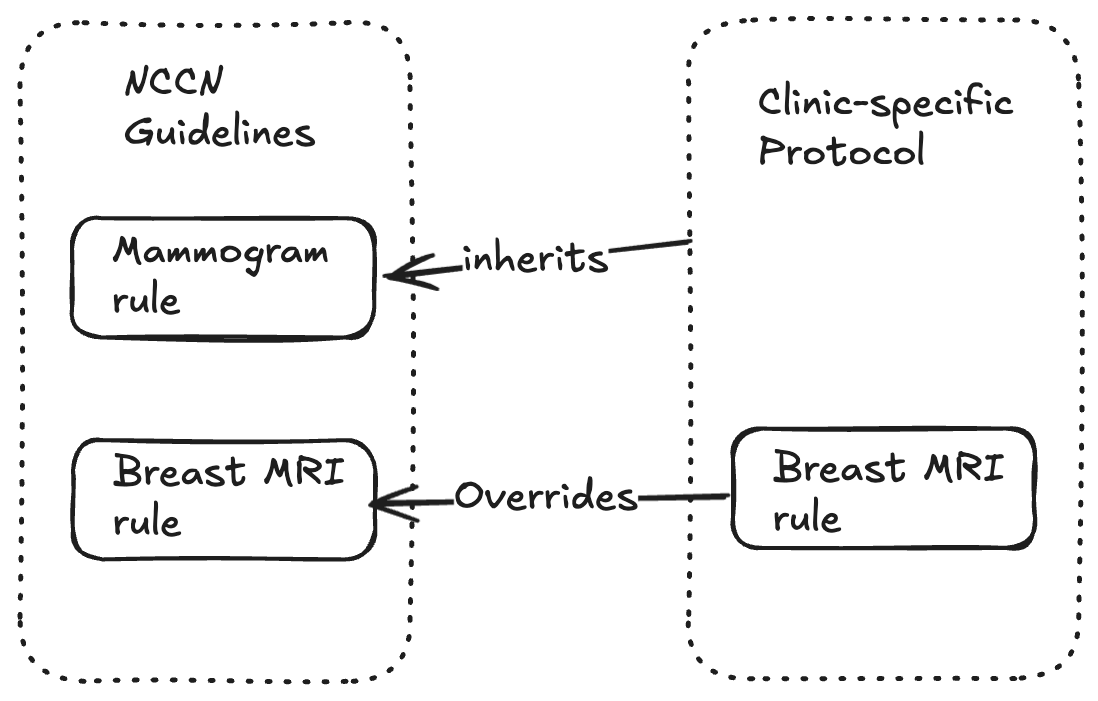}
    \caption{Stacking of Knowledge Bases}
    \label{fig:kb_stacking}
\end{figure}

\paragraph{Customization}
To support customization, we parse each guidelines or clinical protocol document as an independent knowledge base. Then, at inference time, we stack knowledge bases on top of each other, as shown in Figure~\ref{fig:kb_stacking}. As a result, the system recommends items from the set of both knowledge bases. When there is a conflict in the recommendation, the “rule” for what is applicable is pulled from the knowledge base with the highest priority or exposed to the clinician. This prioritization approach makes it possible to implement institutional policies. For example, a hospital such as the University of California, San Francisco (UCSF) may have NCCN Guidelines at the base, overlay on top of these ASCO Guidelines, and have UCSF-specific Guidelines at the top level as highest priority. 

It should be noted that a potential complication in this design is that higher-level rules may unintentionally override partial logic from lower-level rules, causing unexpected side-effects. This is similar to the risks that come with variable overloading in computer programming. We believe that this risk can be minimized by keeping knowledge bases relatively granular (e.g. breast cancer vs. all cancers), but in future versions, detecting and managing such cases could be part of the functionality built into the knowledge base server.

\subsection{Knowledge Base Server}
Given that the LLE architecture assumes a declarative Knowledge Base structure with versioning and namespaces, it becomes possible to support higher level concepts at the infrastructure level rather than requiring the application developer to make single-shot optimizations of their interactions with the LLM. For example, in our use case, a key task is the translation of unstructured and inconsistent health record information to a structure that maps directly to the logic of the workflow (providing a diagnosis, a screening plan, recommendations for a pre-treatment workup, etc). This task can be supported at the Knowledge Base Server (infrastructure) level rather than by the developer of any individual workflow. 

\begin{figure}[htbp]
    \centering
    \includegraphics[width=\linewidth]{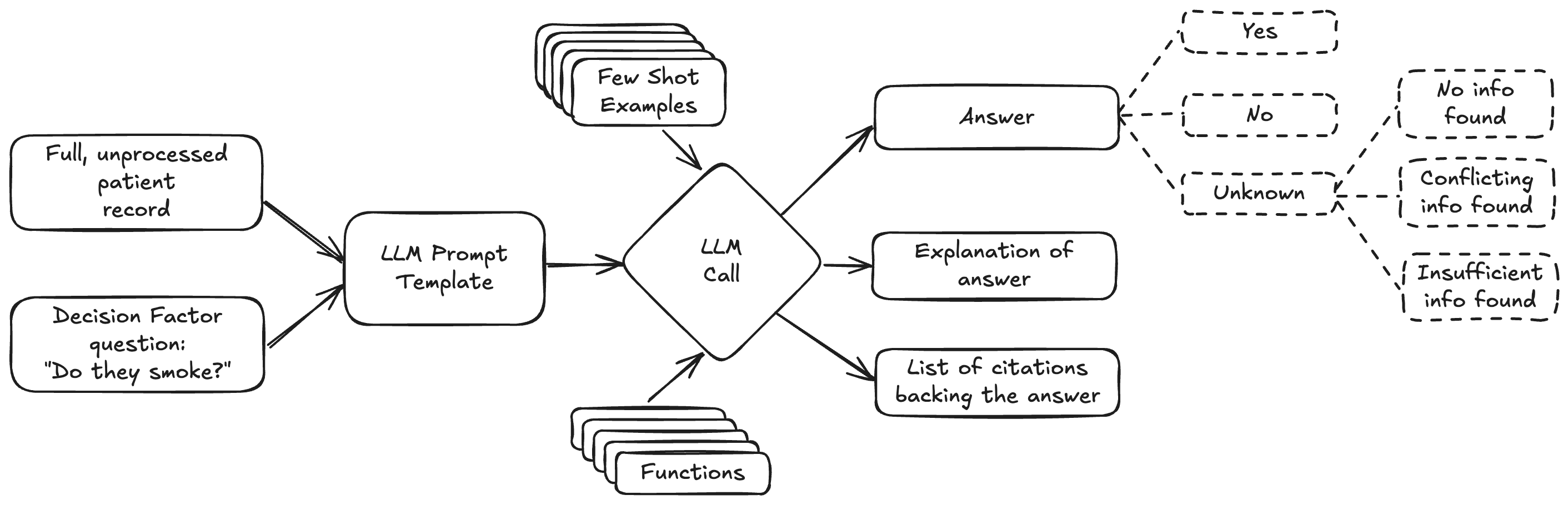}
    \caption{Extraction of Clinical Decision Factors}
    \label{fig:extracting_cdfs}
\end{figure}

\paragraph{Identify and extract clinical decision factors}
Through the knowledge base processing step described above, we have a list of the decision factors that determine the applicability of each recommendation. To address the unknown structure of patient data in the real world, we first collect sources of information we have access to (structured EHR fields, text notes from clinical visits, PDF files, etc.). Next, we extract key concepts from them with another set of LLM requests that are instrumented with function tools to mitigate and solve for common failure modes. Since the factors are extracted as a pair of \texttt{(variable\_name, human\_readable\_question)} tuples, we can use the questions directly with an LLM. Our prompt extracts a “yes”, “no”, or “unknown” answer from the patient record as well as an explanation and a list of cited sentences that the LLM used to reason for this question. The extraction flow is shown in Figure~\ref{fig:extracting_cdfs}.

Depending on the application, the user can review the evaluation of outputs at this stage in the clinical user experience. To build trust and address explainability, the answers are shown in a condensed list where details for the reasoning and cited portions of the patient record can be easily inspected. In case of discrepancies (inaccurate extraction or reasoning) any clinical decision factor can be easily corrected.

This approach has several desirable effects. First, it directly tackles a laborious and error-prone step across many clinical workflows, which is to make sense from health records. Second, when the logic and underlying decision factors are explicitly exposed, errors and ambiguities can be corrected so that errors do not propagate to downstream steps. Third, this approach exposes the same decision factors that an expert would be using in practice (i.e. in the expert’s terminology rather than an abstracted software representation).

The current implementation provides a single output for each factor. Future versions may benefit from the inclusion of a confidence score, which in turn can influence the user interface. For example, a human expert can be prompted to specifically check low confidence outputs, or alternatively trigger a second review when high confidence model outputs are overridden by the user.

\begin{figure}[htbp]
    \centering
    \includegraphics[width=\linewidth]{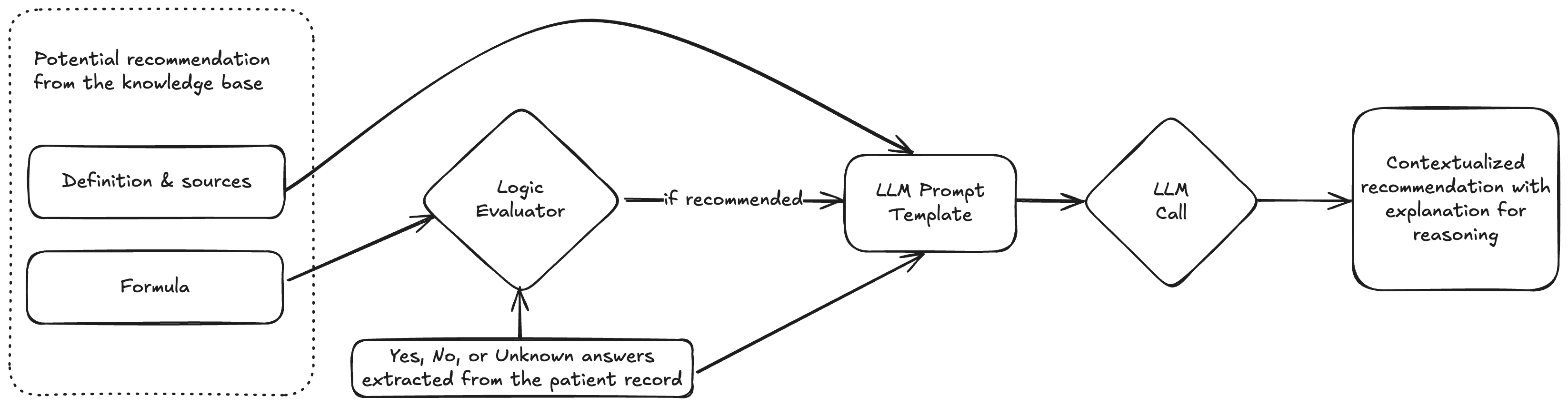}
    \caption{Generation of Recommendations \& Reasoning}
    \label{fig:logic_evaluation}
\end{figure}

\paragraph{Evaluate potential recommendations}
When all clinical decision factors are extracted, the application then proceeds to evaluate against a list of recommendations. The knowledge base processing pipeline extracts “rules” for each potential recommendation. The rules are first-order logic formulas over the list of clinical decision factors we have at hand, and they can be evaluated deterministically, this flow is shown in Figure~\ref{fig:logic_evaluation}. To address explainability concerns, we use an LLM to generate a human-readable summary of why something is recommended based on the patient record, extracted decision factors, and the rule’s first-order logic formula. This explanation, when combined with the detailed reasoning and citations for every underlying question, provides a much greater degree of visibility into the system than what we could accomplish in a purely software-driven rule-based system.

\paragraph{Expert review}
Building our application underpinned by this LLE architecture introduces a modular intervention experience that has two main outcomes:
\begin{enumerate}
    \itemsep0em 
    \item Enabling expert intervention at critical decision points. As a result, this prevents errors from compounding through the system, as the expert is able to make corrections that propagate to the downstream logic. Additionally, the interventions are naturally granular and provide visibility into errors that can easily be corrected through knowledge base updates.
    \item Streamlining the ability to intervene, due to the increased interpretability, explainability, and determinism grounded in the rule-based evaluation of clinical guidelines.
\end{enumerate}
As previously mentioned, this approach enables an application design that is robust to specific system errors or inconsistencies. For example, LLMs are not great at math and date calculations. Knowing this, the clinician is easily able to correct date calculation errors manually without degrading the rest of the system. Furthermore, this enables builders to quickly identify failure modes that reduce the system’s zero-shot ability to arrive at the correct answer. Given the architecture of LLMs (function tools), it is easy to pointedly intervene, patch, test, and deploy fixes as needed.

\section{Real-world application: Identifying pre-treatment workup gaps for patients recently diagnosed with cancer}
After a patient has been diagnosed with cancer and prior to initiation of treatment, they typically need to complete a diagnostic workup (e.g. tests, labs, imaging) to determine their treatment plan. For many cancers, professional guidelines exist that help inform what workup is needed. Unfortunately, access to guideline compliant workup plans in advance of the first oncology visit is rare and during this visit patients discover that they need to wait several more weeks for workup results \cite{13}. Delays in treatment initiation have been associated with worsened outcomes, including reduced survival rates. A systematic review and meta-analysis found that even a four-week delay in cancer treatment is associated with increased mortality for seven cancers. Additionally, every four-week delay in cancer surgery resulted in a 6-8\% increase in the risk of death \cite{14}. \\

Given proper tooling, a clinician would be able to identify relevant workup gaps and ensure they are addressed as early as possible thereby reducing the time to treatment initiation. 

\subsection{Cancer Copilot Overview}
Color’s Cancer Copilot implements the LLE architecture to the problem of workup plan generation. It is a two step human-in-the-loop system that in the first step extracts clinical decision factors required for evaluating workups, as identified by the LLE. In the second step the extracted clinical decision factors are used with the logic evaluator to determine which workups are relevant to a patient, and whether they are completed.

\subsubsection{Step 1: Clinician review of clinical decision factors}
\begin{figure}[htbp]
    \centering
    \caption{Experience for Step 1}

    \begin{subfigure}[b]{\textwidth}
        \centering
        
        \includegraphics[width=0.9\textwidth]{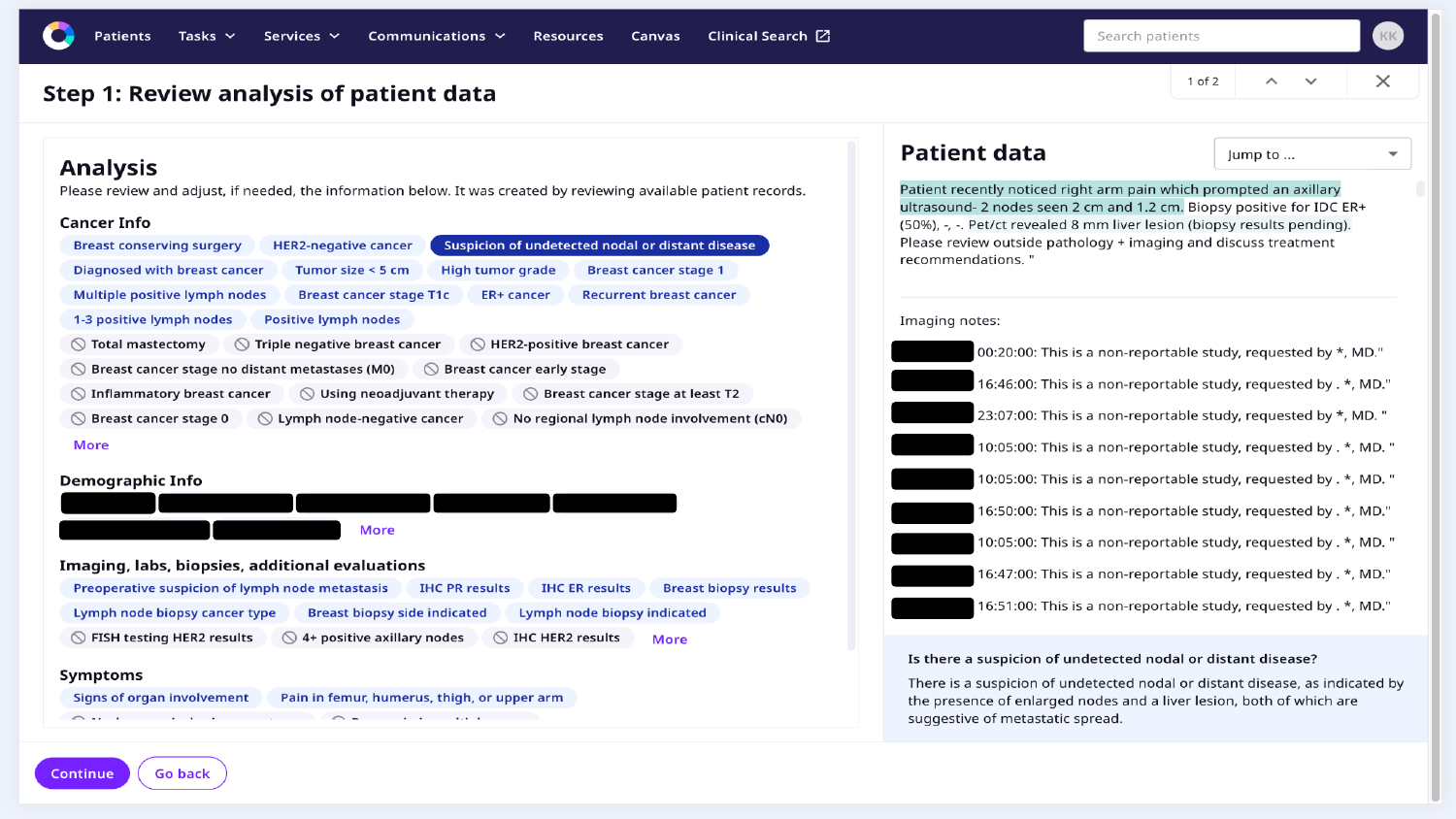}
        \caption{Explanations \& Citations for Clinical Decision Factors}
        \label{fig:inspecting_cdfs}
    \end{subfigure}
    
    \vspace{1em} 

    \begin{subfigure}[b]{\textwidth}
        \centering
        \includegraphics[width=0.9\textwidth]{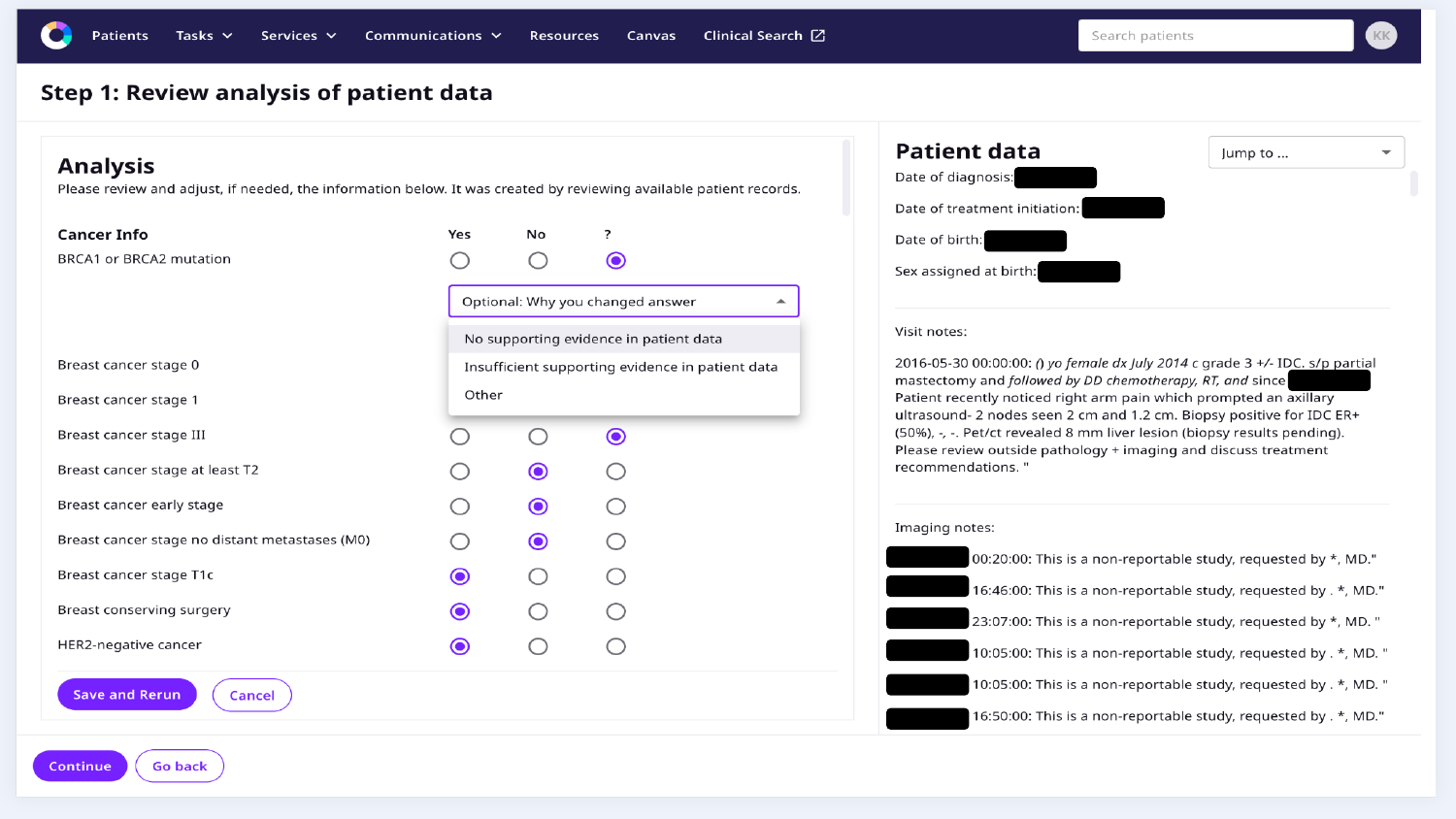}
        \caption{Making Adjustments to Clinical Decision Factors}
        \label{fig:adjusting_cdfs}
    \end{subfigure}

    \label{fig:step1_experience}
\end{figure}

For each patient, the clinician reviews Copilot’s assessment of clinical decision factors. A few examples:
\begin{itemize}
    \itemsep0em 
    \item Does patient have positive lymph nodes (cN+)?
    \item Is there a preoperative suspicion of lymph node metastasis (e.g., through imaging or palpable nodes on exam)?
    \item Are there abnormalities seen on CT or MRI scan that are considered suspicious but inconclusive for metastases?
\end{itemize}
For each clinical decision factor, this includes a yes/no/unknown answer, along with an explanation and list of citations from the patient data explaining the answer. A clinician can adjust answers as needed. This is shown in Figure~\ref{fig:step1_experience}. 

\subsubsection{Step 2: Clinician review of potential workup gaps}
\begin{figure}[htbp]
    \centering
    \caption{Experience for Step 2}

    \begin{subfigure}[b]{\textwidth}
        \centering
        \includegraphics[width=0.9\textwidth]{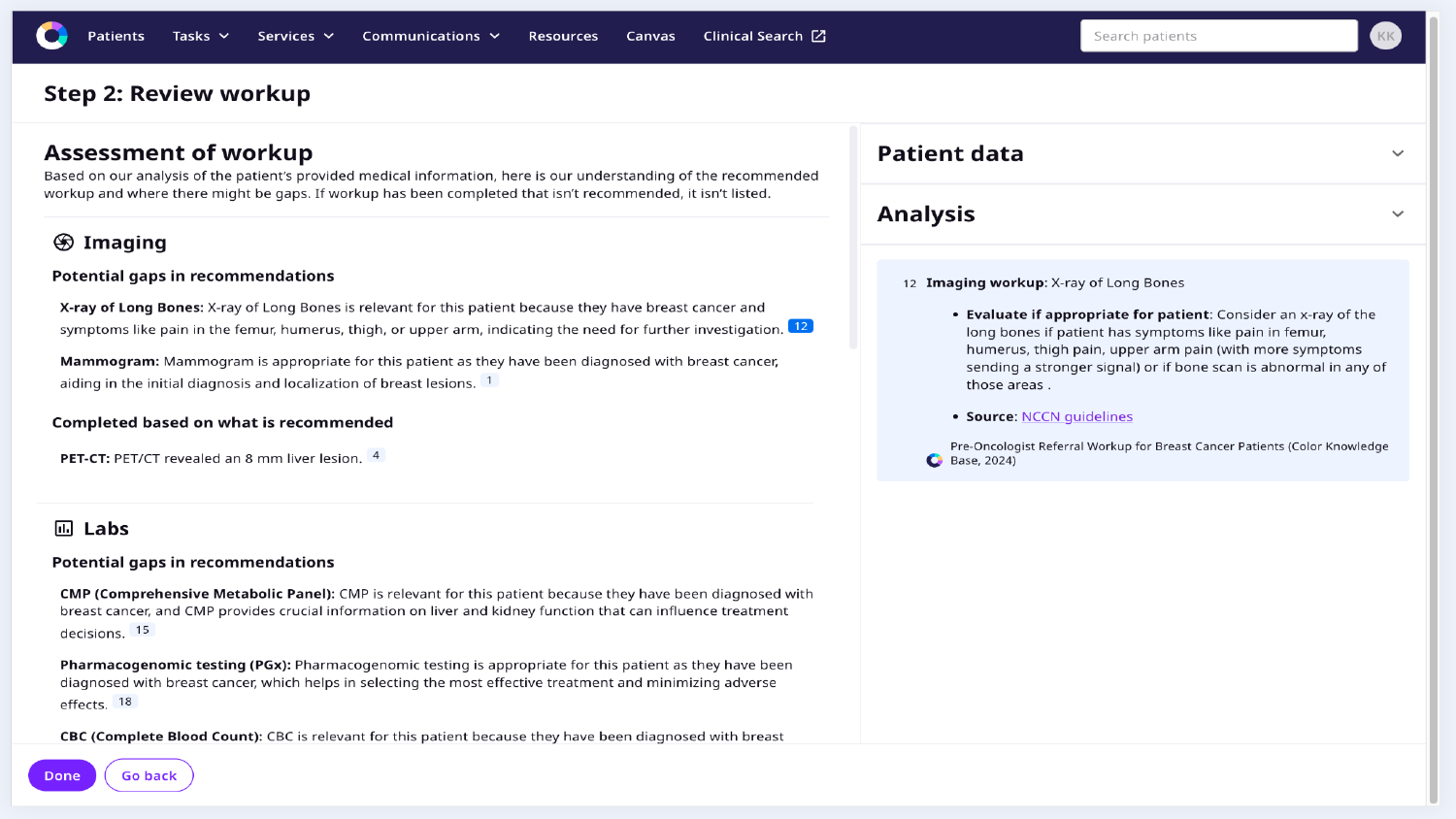}
        \caption{Explanations \& Citations for Recommendations}
        \label{fig:inspecting_citations}
    \end{subfigure}
    
    \vspace{1em} 

    \begin{subfigure}[b]{\textwidth}
        \centering
        \includegraphics[width=0.9\textwidth]{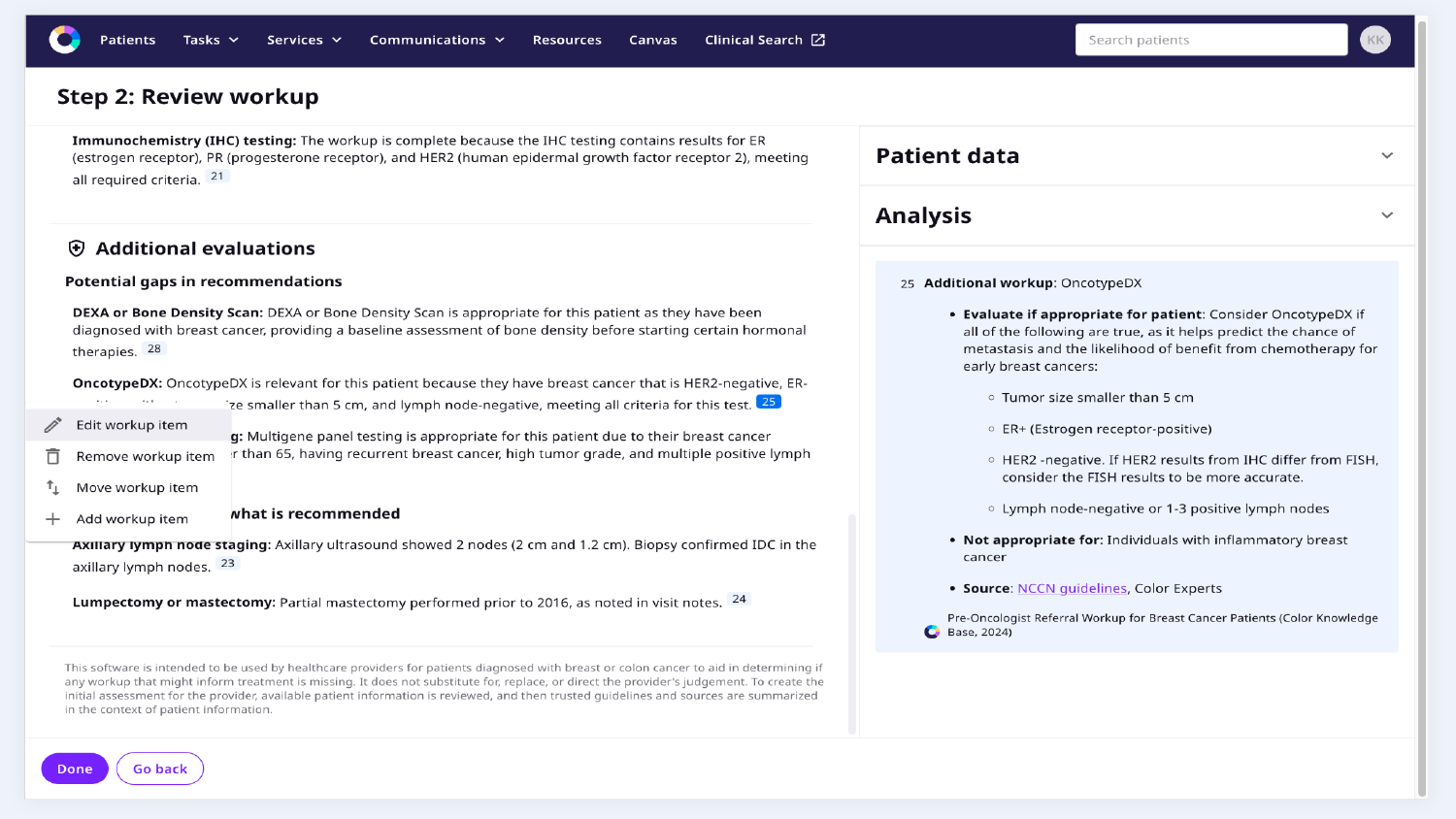}
        \caption{Making Adjustments to Recommendations}
        \label{fig:adjusting_recommendations}
    \end{subfigure}

    \label{fig:step2_experience}
\end{figure}

After reviewing clinical decision factors, the clinician reviews Cancer Copilot’s assessment of recommended workups for the patient. Each recommended workup is categorized as being complete or incomplete (i.e. “a gap”) along with an explanation and related information from the knowledge base for why it was recommended. A clinician can edit, add, move, or remove a workup item as needed. This is shown in Figure~\ref{fig:step2_experience}.

\subsection{Study Design}
Color Health and University of California at San Francisco (UCSF) collaborated on a retrospective study of patients diagnosed with breast and colon cancer. The goal of the study was to determine if the LLE-based tool (“Cancer Copilot”) would enable clinicians to (a) efficiently extract key clinical factors relevant to a newly diagnosed cancer patient and (b) identify which key workup items, as informed by guidelines, were already complete, and which were still required in order to initiate treatment.

UCSF provided Color with 50 de-identified patient cases for breast cancer and 50 for colon cancer. For each patient, we had two sets of records: diagnosis records, which included all available records up to and including the date of diagnosis, and treatment records, which comprised all records up to but not including the date of treatment initiation.

For the study, we processed these records using Cancer Copilot, running 100 patient cases for the diagnosis run type (50 breast and 50 colon) with only the records available up to each patient's date of diagnosis, and 100 patient cases for the treatment run type (50 breast and 50 colon) with all records up to but not including those from when treatment was initiated. A primary care physician employed by Color reviewed the system output and made adjustments as needed.

Performance was evaluated by examining the number of changes made by the clinician to Cancer Copilot’s output in 3 key areas:
\begin{itemize}
    \itemsep0em 
    \item Extracted decision factors
    \item Relevance of recommended workups to the patient
    \item Completeness of relevant workups
\end{itemize}
We additionally recorded the amount of time the clinician spent finalizing the workup plan with Cancer Copilot.

\subsection{Study Results}
\subsubsection{97.9\% of Clinical Decision Factors (Step 1) were not adjusted by clinician}
\begin{figure}[H]
    \centering
    \includegraphics[width=0.8\linewidth]{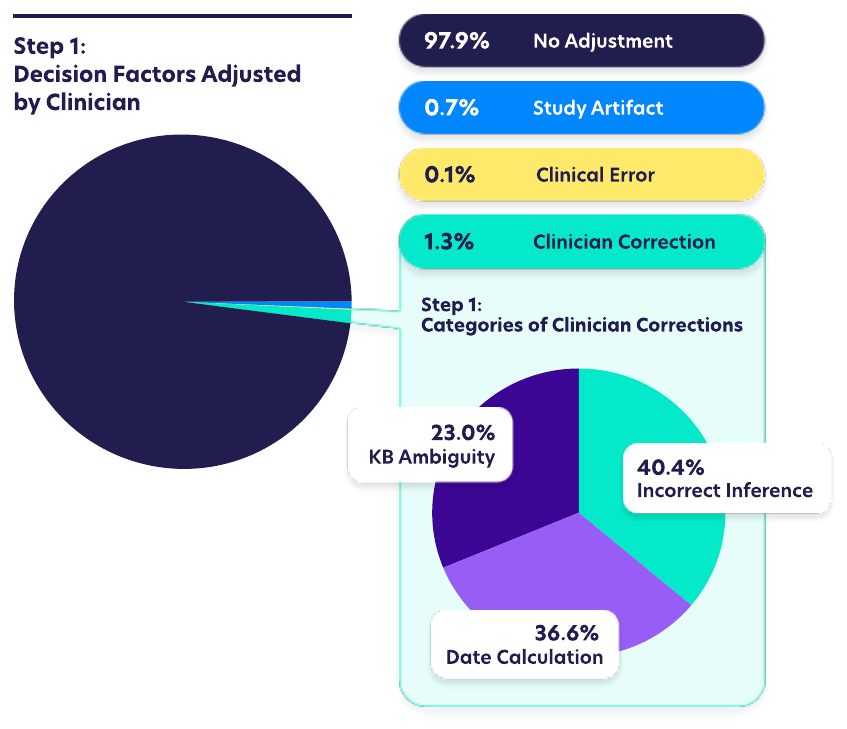}
\end{figure}
Across two runs for 50 breast cancer and 50 colon cancer patients, 12,532 clinical decision factors were extracted by Copilot (8932 for breast, 3600 for colon). The clinician changed 260 outputs, or 2.1\% (172 for breast, 88 for colon). See Table~\ref{tab:granular_step1_changes} for a granular breakdown.

In our post-study analysis, the adjustments made by a clinician 
spanned three high level categories:
\begin{itemize}
    \item Clinician corrections (1.3\%): Valid adjustments made by the clinician. These are the actual errors made by Copilot.
    \item Study artifacts (0.7\%): Manual fixes due to content formatting conditions specific to the study data (e.g. de-identified field placeholders in areas that are non-essential for de-identification). These changes would not be needed outside the context of the study.
    \item Clinician Error (0.1\%): Adjustments erroneously made by the clinician.
\end{itemize}

Table~\ref{tab:step1_clinical_correction_categories} provides a further breakdown of the correct adjustments made by the clinician. 

\begin{table}[htbp] 
\centering
\caption{Clinician Correction Categories for Step 1 (required for 1.3\% of decisions)}
\resizebox{\textwidth}{!}{%
    \begin{tabularx}{\textwidth}{>{\raggedright\arraybackslash}p{1.8cm} | p{2.8cm} | X | X }
        \rowcolor{black} 
        \textcolor{white}{\textbf{Clinician Correction Category}} & 
        \textcolor{white}{\textbf{Description}} & 
        \textcolor{white}{\textbf{Example}} & 
        \textcolor{white}{\textbf{Opportunity for Improvement}} \\ 
        \midrule
        \rowcolor{lightgray}
        \textbf{Incorrect Inference (40.4\%)} & 
        Copilot arrived at an answer unaligned with the clinician’s judgment & 
        \textit{Question:} Is the patient postmenopausal? \newline \newline
        \textit{System Answer:} Yes \newline
        \textit{System Reasoning:} The patient is above the age of 50 \newline \newline
        \textit{Expert Answer:} Unknown \newline
        \textit{Expert Reasoning:} Insufficient supporting evidence in patient data & 
        We expect these cases to be addressed through refinements to the knowledge base logic and LLM temperature settings \& prompts. The highlighted example would be resolved through the incorporation of the logic for determining menopausal status in the knowledge base. Additionally, as foundation models become more competent in medical Q\&A \cite{15}\cite{16}, we expect that these improvements will be reflected in our system.  \\
        \midrule
        \textbf{Date Calculation (36.6\%)} & 
        Copilot didn’t handle factors related to processing dates correctly & 
        \textit{Question:} Is the patient younger than 65 years old? \newline \newline
        \textit{System Answer:} No \newline
        \textit{System Reasoning:} Based on the date of birth, the patient is older than 65 years as of the date of diagnosis \newline \newline
        \textit{Expert Answer:} Yes \newline
        \textit {Expert Reasoning:} Patient is 56 at date of diagnosis & 
        Questions involving date calculations were handled directly by the LLM, a method known to be notoriously error-prone for temporal computations \cite{17}. We aim to mitigate errors in this category by using functions that deterministically evaluate date-related computations and by providing few-shot examples \cite{18}. \\
        \midrule
        \rowcolor{lightgray}
        \textbf{Knowledge Base Ambiguity (23.0\%)} & 
        Copilot’s answer was incorrect because the method for extracting a decision factor, generated in our knowledge curation step, needed greater specificity to be applied in its context & 
        \textit{Question:} Is the patient’s breast cancer stage advanced disease (stage III)? \newline \newline
        \textit{How the question is understood:} Is the patient’s breast cancer stage III? \newline \newline
        \textit{Consequence:} The answer “Yes” would apply to stage IV cases as well, creating a misalignment with how this decision factor should be used. & 
        We haven’t optimized the system that distills guidelines into questions for extracting decision factors and evaluation rules, leaving room for improvement through few-shot examples and better prompting. We expect these refinements, along with advances in reasoning models - seen in the stepwise gains from GPT-4o to OpenAI o1 \cite{12} - to help resolve errors that we see in this category.  \\
        \bottomrule
    \end{tabularx}
}%
\label{tab:step1_clinical_correction_categories}
\end{table}

\subsubsection{95.5\% of Workup Items (Step 2) were not adjusted by clinician}
\begin{figure}[H]
    \centering
    \includegraphics[width=0.8\linewidth]{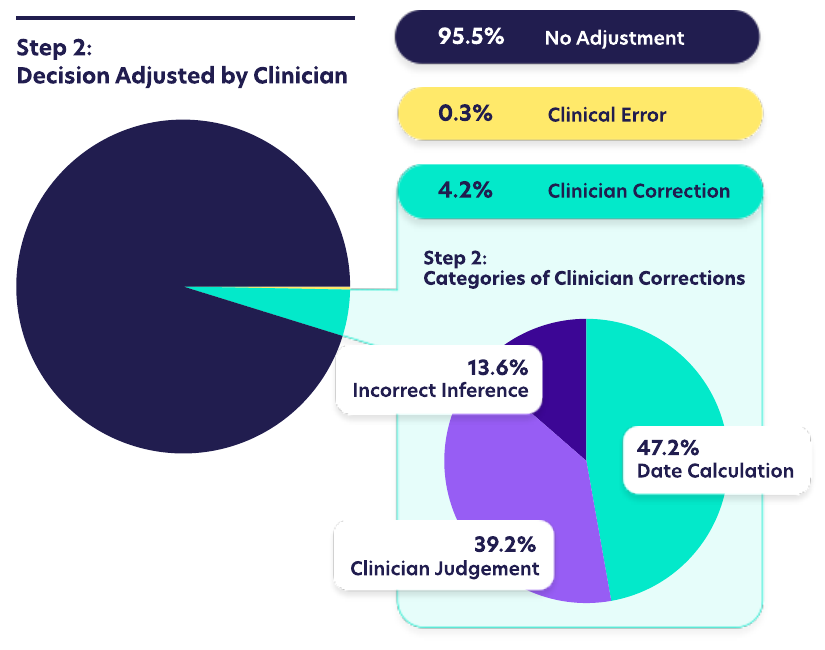}
\end{figure}
Across two runs for 50 breast cancer and 50 colon cancer patients, 2,971 workup items were provided by Copilot (1,423 for breast, 1,548 for colon). The clinician made modifications to 135 workup items, or 4.5\% (51 for breast, 84 for colon). See Table~\ref{tab:granular_step2_changes} for a granular breakdown.

In our post-study analysis, the adjustments made by a clinician spanned two high-level categories: 
\begin{itemize}
    \itemsep0em 
    \item Clinician corrections (4.2\%): Valid adjustments made by the clinician. These are the actual errors made by Copilot.
    \item Clinician error (0.3\%): Adjustments made erroneously by the clinician.
\end{itemize}

Table~\ref{tab:step2_clinical_correction_categories} provides a further breakdown of the correct adjustments made by the clinician. 

\begin{table}[htbp] 
\centering
\caption{Clinician Correction Categories for Step 2 (required for 4.5\% of recommendations)}
\resizebox{\textwidth}{!}{%
    \begin{tabularx}{\textwidth}{>{\raggedright\arraybackslash}p{1.8cm} | p{2cm} | X | p{3.5cm}X }
        \rowcolor{black} 
        \textcolor{white}{\textbf{Clinician Correction Category}} & 
        \textcolor{white}{\textbf{Description}} & 
        \textcolor{white}{\textbf{Example}} & 
        \textcolor{white}{\textbf{Opportunity for Improvement}} \\ 
        \midrule
        \rowcolor{lightgray}
        \textbf{Date Calculation (47.2\%)} & 
        Copilot did not calculate a date correctly & 
        \textit{Original recommendation summary:} "The workup for multigene panel testing is relevant because the patient has been diagnosed with breast cancer and is younger than 65 years old, being 62 years old. This meets the criteria for offering multigene panel testing, as outlined in the clinical guidelines." \newline \newline
        \textit{Clinician updated recommendation summary:} "The workup for multigene panel testing is relevant because the patient has been diagnosed with breast cancer and is younger than 65 years old, being 55 years old. This meets the criteria for offering multigene panel testing, as outlined in the clinical guidelines." &
        Since we use the explanations for the extracted clinical decision factors in the construction of the workup summary, we expect that by improving the accuracy of the extraction, the explanations for the extracted values will improve, which will subsequently show up in the summaries for workups being recommended and resolve errors that fall in this category. \\
        \midrule
        \textbf{Clinical Judgment (39.2\%)} & 
        The clinician made a change that is inconsistent with the KB, but instead informed by their clinical judgment & 
        \textit{Workup:} PET-CT\newline
        \textit{Correction Type:} Removed by clinician\newline\newline
        \textit{Original recommendation summary:} "The PET-CT is relevant because the patient has been diagnosed with breast cancer, is not pregnant, and shows signs of organ involvement, as indicated by renal failure and cardiomegaly."\newline \newline
        \textit {Clinician reason for removing:} "renal failure and cardiomegaly not 2/2 to cancer" & 
        We expect this category of corrections to always be present, as our system generates recommendations strictly based on guidelines. In practice, clinicians combine patient factors with their expertise, sometimes diverging from guidelines based on experience and intuition. This reflects the system’s intended role—to support decision-making while recognizing that clinical judgment will sometimes override guideline-based recommendations.
        \\
        \midrule
        \rowcolor{lightgray}
        \textbf{Incorrect Inference (13.6\%)} & 
        Copilot’s assessment that a workup was relevant to the patient, or whether it had been already completed, was incorrect & 
        \textit{Workup:} PET-CT\newline
        \textit{Correction Type:} Removed by clinician\newline\newline
        \textit{Original recommendation summary:} "The patient is suspected to have metastatic colon cancer with suspicious findings on CT, making PET-CT relevant for further evaluation."\newline \newline
        \textit {Clinician reason for removing:} "CT didn't show evidence of metastatic disease" & 
        The recommendations that are generated in this step are deterministically evaluated based on the clinical decision factors extracted in the previous step. We expect that many of the adjustments in this category would be resolved by improving the accuracy of that extraction. \\
        \bottomrule
    \end{tabularx}
}%
\label{tab:step2_clinical_correction_categories}
\end{table}

\subsubsection{Time Spent: Non-specialists unfamiliar with patient cases take less than 7.5 minutes to finalize recommendations}

Along with assessing the system's capability to generate guideline-concordant recommendations, we also measured the time spent by a non-specialist physician, unfamiliar with the patient cases, at each step of the workflow. For colon cancer, the median time spent by the clinician to approve the extracted decision factors (step 1) was 2.3 minutes, followed by a median time of 2.1 minutes to finalize the workup recommendations (step 2). For breast cancer, these times were 5.2 \& 2.1 minutes, respectively.

It should be noted that these times should be considered directional, given the retrospective study conditions. Even directionally, though, these are a considerable improvement over the hours anecdotally shared by physicians and their teams that are spent on reviewing new patient cases in the context of guidelines. 

\newpage
\section{Conclusions}
\subsection{LLE in Practice - Color Cancer Copilot}
Our study provides a promising demonstration of the potential for Large Language Experts (LLEs) to measurably improve the delivery of high-quality, guideline-concordant cancer care. By bringing together the strengths of large language models (LLMs), structured knowledge bases, and a streamlined experience, the Cancer Copilot enables clinicians to efficiently review patient records and identify workup gaps, while maintaining a very high level of accuracy.

Across the real-world breast \& colon cancer cases that were used to test the system, the correction rate on Copilot’s recommendations, measured through changes by the clinician, fell below 5\% (2.1\% on clinical decision factors extracted, 4.5\% on workups recommended).  Necessitating such infrequent interventions by a clinician indicates the effectiveness of the LLE architecture in faithfully encoding and applying complex clinical logic to real-world patient data. 

Even at this initial performance level, the Copilot represents a tool with the potential to meaningfully reduce busywork and enhance clinical quality. The <5\% correction rate falls short of what would be needed for purely autonomous use. However, the limited human input required to reach clinical-level accuracy took less than 7.5 minutes per patient case on average - a substantial improvement over the hours often spent reviewing records and guidelines to determine appropriate workup outside of the study setting. 

\subsection{LLE Architecture}
Compared to pure machine learning approaches, the LLE architecture proved highly effective and efficient to implement in practice. The initial knowledge base development required a focused but manageable effort, with the LLM-assisted translation from natural language guidelines to structured logic. The resulting system achieved high baseline performance without the need for extensive model fine-tuning or large labeled training datasets. 

More importantly, when issues did arise, the transparent rule definitions enabled rapid diagnosis and resolution. The limited set of failure modes seen in the study, such as challenges with temporal math or occasional spurious logical leaps, point to clear opportunities for further system refinement. For example, key calculations like age could be delegated to specialized sub-functions, and the knowledge base logic could undergo additional adversarial testing to identify and tighten overly permissive rules.

Looking beyond the scope of this initial application, the LLE architecture may offer a powerful new approach to implementing guideline-based care more broadly. Many healthcare domains involve intricate-but-standardized workflows that have to interface with an inconsistent and largely unstructured data reality. Potential use cases could include clinical trial matching, emergency department triage, insurance pre-authorizations, quality measures, and the scaling of access to scarce expert knowledge. 

The ability of the LLE approach to smoothly integrate centrally-curated guidelines with site-specific customizations makes it appealing as a framework that can balance consistency with flexibility. As comfort and capabilities grow, we envision an ecosystem of knowledge bases published by professional societies, adapted and validated by individual institutions, and made accessible through purpose-built clinical applications.

Ultimately, our results suggest that thoughtful combination of language models with more traditional knowledge representational structures can enable a new level of capable, reliable, and user-friendly systems. With further validation and hardening, this approach could help transform the way that many forms of specialized expertise are disseminated and applied to real-world decision-making.

\clearpage

\appendix
\section{Study Results}
\label{appendix:study_results}

\subsection{Clinical Decision Factor Adjustments}
\begin{table}[h]
    \centering
    \caption{Changes in Clinical Decision Factor Answers}
    \label{tab:granular_step1_changes}
    \renewcommand{\arraystretch}{1.2}
    \begin{tabular}{l l S S S}  
        \toprule
        \textbf{Disease Area} & \textbf{Run Type} & \textbf{Factors Extracted} & \textbf{Factors Adjusted} & \textbf{Adjustment Percentage (\%)} \\
        \midrule
        \multirow{3}{*}{\textbf{Breast}}  
            & Diagnosis  & 4466 & 84  & 1.88 \\
            & Treatment  & 4466 & 88  & 1.97 \\
            & \textbf{Combined}  & 8932 & 172 & \hspace{0.4em}\textbf{1.93} \\
        \midrule
        \multirow{3}{*}{\textbf{Colon}}  
            & Diagnosis  & 1800 & 59  & 3.28 \\
            & Treatment  & 1800 & 29  & 1.61 \\
            & \textbf{Combined}  & 3600 & 88 & \hspace{0.4em}\textbf{2.44} \\
        \midrule
        \multirow{3}{*}{\textbf{Total}}  
            & Diagnosis  & 6266 & 143 & 2.28 \\
            & Treatment  & 6266 & 117 & 1.87 \\
            & \textbf{Combined}  & 12532 & 260 & \hspace{0.4em}\textbf{2.07} \\
        \bottomrule
    \end{tabular}
\end{table}

\subsection{Workup Item Adjustments}
\begin{table}[h]
    \centering
    \caption{Changes in Workup Recommendations}
    \label{tab:granular_step2_changes}
    \renewcommand{\arraystretch}{1.2}
    \begin{tabular}{l l S S S}  
        \toprule
        \textbf{Disease Area} & \textbf{Run Type} & \textbf{Recommendations} & \textbf{Adjustments} & \textbf{Adjustment Percentage (\%)} \\
        \midrule
        \multirow{3}{*}{\textbf{Breast}}  
            & Diagnosis  & 699  & 17  & 2.43 \\
            & Treatment  & 724  & 34  & 4.70 \\
            & \textbf{Combined}  & 1423 & 51  & \hspace{0.4em}\textbf{3.58} \\
        \midrule
        \multirow{3}{*}{\textbf{Colon}}  
            & Diagnosis  & 737  & 32  & 4.34 \\
            & Treatment  & 811  & 52  & 6.41 \\
            & \textbf{Combined}  & 1548 & 84  & \hspace{0.4em}\textbf{5.43} \\
        \midrule
        \multirow{3}{*}{\textbf{Total}}  
            & Diagnosis  & 1436 & 49  & 3.41 \\
            & Treatment  & 1535 & 86  & 5.60 \\
            & \textbf{Combined}  & 2971 & 135 & \hspace{0.4em}\textbf{4.54} \\
        \bottomrule
    \end{tabular}
\end{table}

\end{document}